# Inferring deterministic causal relations

Povilas Daniušis[1,2], Dominik Janzing[1], Joris Mooij[1], Jakob Zscheischler[1],
Bastian Steudel[3], Kun Zhang[1], Bernhard Schölkopf[1]

[1]Max Planck Institute for Biological Cybernetics, Spemannstr. 38, 72076 Tübingen, Germany
[2]Vilnius University, Faculty of Mathematics and Informatics, Naugarduko St. 24, LT-03225 Vilnius, Lithuania
[3]Max Planck Institute for Mathematics in the Sciences, Inselstrasse 22, 04103 Leipzig, Germany

## Abstract

We consider two variables that are related to each other by an invertible function. While it has previously been shown that the dependence structure of the noise can provide hints to determine which of the two variables is the cause, we presently show that even in the deterministic (noise-free) case, there are asymmetries that can be exploited for causal inference. Our method is based on the idea that if the function and the probability density of the cause are chosen independently, then the distribution of the effect will, in a certain sense, depend on the function. We provide a theoretical analysis of this method, showing that it also works in the low noise regime, and link it to information geometry. We report strong empirical results on various real-world data sets from different domains.

## 1 Introduction

Inferring causal relations among random variables via observing their statistical dependences is a challenging task. For causal directed acyclic graphs (DAGs) of $n > 2$ random variables, causal learning by analyzing conditional independences is among the most popular approaches [Spirtes et al., 1993, Pearl, 2000]. The fundamental limitation of this method, however, is its inability to distinguish between Markov-equivalent DAGs that impose the same set of independences. In particular, the method fails for inferring whether $X \to Y$ ("$X$ causes $Y$") or $Y \to X$ ("$Y$ causes $X$") from the joint distribution $P(X, Y)$. To encourage the development of novel methods for this challenging problem, the NIPS 2008 workshop *Causality - Objectives and Assessment* contained `CauseEffectPairs` as one of its tasks [Mooij and Janzing, 2010].

Two related methods achieved quite reasonable results for this task. (1) Hoyer et al. [2009] introduce *additive noise models*, which we explain here only for the special case of two variables: Assuming that the effect $Y$ is a function of the cause $X$ up to an additive noise term $E$ that is statistically independent of $X$, i.e.,

$$Y = f(X) + E \quad \text{with } E \perp\!\!\!\perp X \,,$$

there will be no such additive noise model model in the reverse direction $Y \to X$ in the generic case. (2) A generalization of this model is the *post-nonlinear model* of Zhang and Hyvärinen [2009] where an additional nonlinear transformation $h$ of the effect is allowed:

$$Y = h\bigl(f(X) + E\bigr) \quad \text{with } E \perp\!\!\!\perp X \,.$$

The statistical *independence* of the noise is the decisive assumption here, since a model with *uncorrelated* noise can always be found.

From this point of view it seems an even harder challenge to distinguish between $X \to Y$ and $Y \to X$ if the relation is deterministic:

$$Y = f(X) \,.$$

If $f$ is not invertible, one will clearly consider the task as solvable, so in this paper we focus on the invertible case. Without loss of generality, we assume that $f$ is monotonically increasing. The idea is the following:

**Postulate 1 (Indep. of input and function)**
*If $X \to Y$, the distribution of $X$ and the function $f$ mapping $X$ to $Y$ are independent since they correspond to independent mechanisms of nature.*

While this postulate is informal in the sense that we do not specify what we mean by independence, we point out that Lemeire and Dirkx [2006] and Janzing and Schölkopf [2008] give a formalization in terms of *algorithmic* independence. Following their ideas, we should postulate that the shortest description of $P(X, Y)$ is given by separate descriptions of $P(X)$ and $f$ (in the sense of Kolmogorov complexity). However, since Kolmogorov complexity is uncomputable, practical implementations must rely on other notions of dependence. Example 1 and Figure 1 illustrate what kind of dependences we expect if we consider the causal relation in the *wrong* direction.

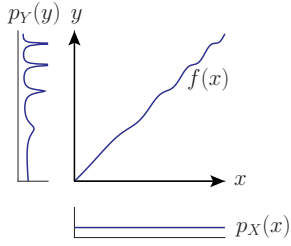

Figure 1: Illustration corresponding to Example 1. Note that the output density $p_Y(y)$ is strongly peaked where $g'$ is large (or equivalently, where $f'$ is small).

**Example 1 (uniform density as input)**
Let $p_X(x) = 1$ be the uniform density on $[0,1]$ and $f$ be a diffeomorphism of $[0,1]$ with $f(0) = 0$ and $f(1) = 1$ and inverse $g := f^{-1}$. Then the distribution of $Y = f(X)$ is given by

$$p_Y(y) = g'(y) = \frac{1}{f'(f^{-1}(y))}. \qquad (1)$$

To see why the causal hypothesis $Y \to X$ is unlikely, observe that it would require mutual adjustments between the hypothetical input density $p_Y(y)$ and the hypothetical mechanism mapping $Y$ to $X$, because the peaks of $p(y)$ coincide with the points of steep slope of $g$.[1]

The following section shows that even for general input density $p(x)$, the peaks of $p(y)$ correlate with the points of $g$ with steep slope. The decisive condition is that the shapes of $p_X$ and $f$ are sufficiently uncorrelated in a sense to be defined below.

## 2 Independence of $f$ and $p_X$ in terms of information geometry

Before we develop the general theory, we start with a generalization of Example 1 by dropping the assumption that the input distribution $p(x)$ is uniform. Let $f$ be as in Example 1 and let $p(x)$ be an arbitrary distribution on $[0,1]$, but assume additionally that

$$\int_0^1 \log f'(x) p(x) dx = \int_0^1 \log f'(x) dx. \qquad (2)$$

---
[1]We write $p(x)$ and $p(y)$ instead of $p_X(x)$ and $p_Y(y)$ whenever this causes no confusion.

To justify why (2) should hold (approximately) for $X \to Y$ is the most delicate part of our theory. Here we give the following interpretation: consider $p(x)$ and $\log f'(x)$ as two random variables on the interval $[0,1]$. Then their covariance with respect to the uniform distribution on $[0,1]$ reads

$$\mathbb{C}\mathrm{ov}\big(\log f'(x), p(x)\big)$$
$$= \int \log f'(x) p(x) dx - \int \log f'(x) dx \int p(x) dx$$
$$= \int \log f'(x) p(x) dx - \int \log f'(x) dx.$$

Hence, (2) expresses the assumption that peaks of $p_X$ do not correlate with regions of large slope of $f$.

Remarkably, we obtain for the backward direction (with $g := f^{-1}$) a positive correlation between $p(y)$ and $\log g'(y)$. To see this, note that (2) is equivalent to

$$\int_0^1 \log g'(y) p(y) dy = \int_0^1 \log g'(y) g'(y) dy.$$

Introducing the uniform distribution $v = U(0,1)$ on $[0,1]$ and observing that $g' = p(y)$, we obtain

$$\mathbb{C}\mathrm{ov}\big(\log g'(y), p(y)\big) = \int (g'(y) - 1) \log g'(y) dy$$
$$= D(g' \,||\, v) + D(v \,||\, g') \geq 0,$$

where $D(\cdot \,||\, \cdot)$ denotes the relative entropy distance.[2]

### 2.1 Reformulation in information space

To show that condition (2) can also be phrased in terms of information geometry, we first mention the following result which can be checked by straightforward computation:

**Lemma 1 (orthogonality in inform. space)**
Let $q, r, s$ be three probability densities. Then

$$\int q(x) \log \frac{r(x)}{s(x)} dx = \int r(x) \log \frac{r(x)}{s(x)} dx \qquad (3)$$

is equivalent to the following additivity of relative entropy distances:

$$D(q \,||\, r) + D(r \,||\, s) = D(q \,||\, s). \qquad (4)$$

---
[2]The *relative entropy distance* or *Kullback-Leibler divergence* between two probability densities $p$ and $q$ is defined by $D(p \,||\, q) := \int p(x) \log \frac{p(x)}{q(x)} dx$. In this paper, we use two elementary properties: (i) positivity, i.e., $D(p \,||\, q) \geq 0$, and (ii) invariance under bijective mappings, i.e., $D(p_f \,||\, q_f) = D(p \,||\, q)$ for any bijection $f$, where $p_f$ and $q_f$ are the images of $p$ and $q$ under $f$ on the corresponding variable.

Condition (4) is called "orthogonality" between $\overrightarrow{qr}$ and $\overrightarrow{rs}$ [Amari, 2001]. Here we use a uniform distribution as a reference measure: Let $u(x)$ and $v(y)$ denote the uniform distributions for $X$ and $Y$, respectively and $u_f$ and $v_g$ be their images under $f$ and $g$:

$$u_f(y) := u(g(y))g'(y) = g'(y)$$

and

$$v_g(x) := v(f(x))f'(x) = f'(x).$$

Applying Lemma 1, we conclude that condition (2) is equivalent to

$$D(p_X \,||\, v_g) = D(p_X \,||\, u) + D(u \,||\, v_g), \quad (5)$$

i.e., orthogonality between $\overrightarrow{p_X u}$ and $\overrightarrow{uv_g}$. Since relative entropy is conserved under bijective maps, (5) in turn is equivalent to

$$D(p_Y \,||\, v) = D(p_X \,||\, u) + D(u_f \,||\, v). \quad (6)$$

Since $D(p_X \,||\, u)$ depends only on the structure of $p_X$ and $D(u_f \,||\, v)$ depends on the structure of $f$, it seems reasonable to postulate that the vectors $\overrightarrow{p_X u}$ and $\overrightarrow{uv_g}$ are typically (close to) orthogonal. We interpret (6) by saying that the amount of irregularities of $p_Y$ is given by the irregularities of $p_X$ plus those of $f$.

### 2.2 Exponential families of reference measures

The uniform distribution may not always be appropriate as reference measure, for example if the distributions do not have compact support (e.g., for Gaussians). We therefore introduce *exponential manifolds* of densities as reference. A natural example that we will use later is the family of all Gaussian distributions on $\mathbb{R}^d$.

The projection of a density $r$ onto an exponential manifold $\mathcal{E}$ is defined[3] by $\mathrm{argmin}_{q \in \mathcal{E}} D(r \,||\, q)$, and we write $D(r \,||\, \mathcal{E}) := \min_{q \in \mathcal{E}} D(r \,||\, q)$. An advantage of considering exponential manifolds is that these projections are unique.

### 2.3 Causal inference in information space

The following assumption generalizes (6) and provides the particular formalization of Postulate 1 that we will study in this paper:

**Postulate 2 (independence and rel. entropy)**
Let $\mathcal{E}_X$ and $\mathcal{E}_Y$ define exponential families of "smooth" reference distributions for $X$ and $Y$, respectively. Let

---
[3] We assume that the exponential families are chosen such that projections always exist, e.g., as manifolds of Gaussians.

$u$ denote the projection of $p_X$ onto $\mathcal{E}_X$ and $u_f$ its image under $f$. If $X \to Y$, then

$$D(p_Y \,||\, \mathcal{E}_Y) = D(p_X \,||\, \mathcal{E}_X) + D(u_f \,||\, \mathcal{E}_Y). \quad (7)$$

In this more general setting, we measure irregularities by the "distance" to an exponential family (instead of having one unique reference measure), but still interpret (7) similar to (6): the amount of irregularities of the effect is given by the irregularities of the cause plus those of the function. For this interpretation it is important that $\mathcal{E}_X$ and $\mathcal{E}_Y$ are manifolds of low dimensions to avoid that too much information gets lost in the projection.

An immediate consequence of Postulate 2 and the positivity of relative entropy is that if $X \to Y$,

$$C_{X \to Y} := D(p_X \,||\, \mathcal{E}_X) - D(p_Y \,||\, \mathcal{E}_Y) \leq 0\,.$$

On the other hand, if $Y \to X$, we obtain

$$C_{Y \to X} := D(p_Y \,||\, \mathcal{E}_Y) - D(p_X \,||\, \mathcal{E}_X) \leq 0\,.$$

Therefore, except in cases where $D(u_f \,||\, \mathcal{E}_Y) = 0$ (i.e., if the function is "too simple"), we can simply infer the causal direction by looking at the sign of $C_{X \to Y} = -C_{Y \to X}$.

Thus, we propose the following causal inference method, which we call "Information Geometric Causal Inference (IGCI)":

**Causal Inference method (IGCI)**: *Given $C_{X \to Y}$, infer that $X$ causes $Y$ if $C_{X \to Y} < 0$, or that $Y$ causes $X$ if $C_{X \to Y} > 0$.*

We will see in Section 4 that this method can even be justified in the low noise regime.

## 3 Special cases and estimators

If $f$ is a diffeomorphism between submanifolds of $\mathbb{R}^d$ we can make $C_{X \to Y}$ more explicit in order to obtain a relationship that will be useful in practice. Let $u$ and $v$ be the projections of $p_X$ and $p_Y$ on $\mathcal{E}_X$ and $\mathcal{E}_Y$, respectively. Then:

$$D(p_X \,||\, \mathcal{E}_X) = D(p_X \,||\, u) = -S(p_X) - \int p(x) \log u(x)dx$$
$$= -S(p_X) + S(u),$$

where $S(\cdot)$ denotes the differential entropy. The last equation is not completely obvious, so we will explain it in more detail. First, assume that $\mathcal{E}_X$ contains the uniform distribution $u_0$. Because $u$ is the projection of $p_X$ onto $\mathcal{E}_X$, $\overrightarrow{p_X u}$ and $\overrightarrow{uu_0}$ are orthogonal [Amari, 2001]. Using Lemma 1, we obtain

$-\int p(x) \log u(x) dx = S(u)$. In general, one can imagine that elements of $\mathcal{E}_X$ come arbitrarily close to the uniform distribution $u_0$, and the identity follows by taking an appropriate limit.

As the entropy of $f(X)$ is $S(p_{f(X)}) = S(p_X) + \int p_X(x) \log |\det \nabla f(x)| \, dx$, where $\nabla f$ is the Jacobian of $f$, we have

$$C_{X \to Y} = D(p_X \,\|\, \mathcal{E}_X) - D(p_Y \,\|\, \mathcal{E}_Y)$$
$$= \big(S(u) - S(p_X)\big) - \big(S(v) - S(p_Y)\big) \quad (8)$$
$$= S(u) - S(v) + \int \log |\det \nabla f(x)| \, p(x) dx. \quad (9)$$

In the next subsections, we will consider several special cases in detail, which will provide more intuition for the interpretation of the exponential manifolds of reference measures.

### 3.1 Uniform reference measure on $[0,1]$

Let $\mathcal{E}_X$ and $\mathcal{E}_Y$ contain only the uniform distribution on $[0,1]$. Then the entropies of $u$ and $v$ coincide and (9) implies that for every diffeomorphism $f$ on $[0,1]$ we obtain

$$C_{X \to Y}^U = \int_0^1 \log |f'(x)| \, p(x) dx. \quad (10)$$

Note that the proposed causal inference method now is closely related to the assumption made at the beginning of Section 2.

### 3.2 Gaussian reference measure on $\mathbb{R}^d$

Suppose that both $X$ and $Y$ are $d$-dimensional real random vectors, and $f$ is a diffeomorphism $\mathbb{R}^d \to \mathbb{R}^d$. Let both $\mathcal{E}_X$ and $\mathcal{E}_Y$ be the manifold of $d$-dimensional Gaussian distributions. The projections $u$ and $v$ are again Gaussians with the same mean and variance as $X$ and $Y$. The difference of the entropies of $u$ and $v$ thus reads $\frac{1}{2} \log(\det \Sigma_X / \det \Sigma_Y)$. Then we can easily find $C_{X \to Y}^G$ based on (9).

We now focus on the case $d = 1$ in order to compare with Subsection 3.1. Suppose that $X$ and $Y$ both take values on $[0,1]$. We can then use either criterion (10) or $C_{X \to Y}^G$ for causal inference. The difference between these criteria is easily seen to be $C_{X \to Y}^G - C_{X \to Y}^U = \log \sigma_X - \log \sigma_Y$, where $\sigma_X^2$ and $\sigma_Y^2$ denote the variances of $X$ and $Y$, respectively. Consequently, if $\sigma_X^2 = \sigma_Y^2$, the choice of the reference measures does not make any difference. However, depending on the shape of the distributions of $X$ and $Y$ and possible outlier effects, $\sigma_X^2$ and $\sigma_Y^2$ may be different. This means that $C_{X \to Y}^U$ and $C_{X \to Y}^G$ may have different signs (and consequently, would imply a different causal direction), although these cases were surprisingly rare in our experiments.

### 3.3 Isotropic Gaussians as reference on $\mathbb{R}^d$

We will now show that a method described in [Janzing et al., 2009] can be rephrased as a special case of ours. Let $p_X$ and $p_Y$ be multivariate Gaussians in $\mathbb{R}^d$ and $X$ and $Y$ be related by $Y = AX$ where $A$ is an invertible $d \times d$-matrix.[4] Let $\mathcal{E}_X$ and $\mathcal{E}_Y$ be the manifold of isotropic Gaussians, i.e., those whose covariance matrices are multiples of the identity. For an arbitrary $d \times d$ matrix $B$ let $\tau(B) = tr(B)/d$ denote the renormalized trace. Then, $u$ and $v$ have the same mean as $X$ and $Y$ and their covariance matrices read $\tau(\Sigma_X)\mathbf{I}$ and $\tau(\Sigma_Y)\mathbf{I}$.

Janzing et al. [2009] argue that for any given $A$, choosing $\Sigma_X$ according to a rotation invariant prior ensures that the following expression is close to zero with high probability:

$$\Delta := \log \tau(A \Sigma_X A^T) - \log \tau(AA^T) - \log \tau(\Sigma_X). \quad (11)$$

This is proved via a concentration of measure phenomenon. Based on this statement, it is proposed to prefer the causal direction for which $\Delta$ is closer to zero.

Rephrasing the results of Section 2 in [Janzing et al., 2009] into our language, they show that

$$D(p_Y \,\|\, \mathcal{E}_Y) = D(p_X \,\|\, \mathcal{E}_X) + D(u \,\|\, A^{-1} \mathcal{E}_Y) + \frac{d}{2}\Delta.$$

can be phrased in terms of traces and determinants, i.e.,

$$D(p_X \,\|\, \mathcal{E}_X) = \frac{1}{2}\big(d \log \tau(\Sigma_X) - \log \det(\Sigma_X)\big).$$

Hence, condition (7) is equivalent to $\Delta = 0$ and the restriction of the inference method in [Janzing et al., 2009] to deterministic linear models is a special case of our method. Remarkably, for this case, Postulate 2 thus gets an additional justification via a probabilistic scenario where $f$ is fixed and $p_X$ is chosen randomly from a prior satisfying a certain symmetry condition.

Note that the method described in Subsection 3.2 makes use of the non-linearities of $f$, while it removes the information that is contained in $\Sigma_X$, since the set of *all* Gaussians is used as reference. For relations that are close to the linear case, one thus looses the essential information. In such cases, therefore, one has to apply the method above: taking isotropic Gaussians as reference ensures that only the information that describes the joint (overall) scaling is lost.

---

[4]Janzing et al. [2009] also consider the case $Y = AX + E$ for some noise $E$, but we restrict the attention to the deterministic one.

### 3.4 Non-uniform reference measure on finite sets

The intuitive explanation of the identifiability of cause and effect used the fact that regions of high density of the effect correlate with regions of high slope of the inverse function. Remarkably, our method is in principle also applicable for bijections between *finite* probability spaces, provided that we ensure that $D(u_f \| \mathcal{E}_Y) > 0$, i.e., $\mathcal{E}_X$ and $\mathcal{E}_Y$ must not consist of the uniform distribution only. We omit the details but only give a brief sketch of a special case here.[5]

Assume that both $X$ and $Y$ take values in $\{1, \ldots, k\}$ and $p_X$ and $p_Y$ are probability mass functions with $p_Y(y) = p_X(g(y))$. Let $\mathcal{E}_X$ and $\mathcal{E}_Y$ be the two-parametric manifold of distributions of "discrete Gaussians" with

$$u(x \,|\, \mu, \sigma) \propto e^{-\frac{(x-\mu)^2}{2\sigma^2}} \,,$$

where $\mu \in \mathbb{R}$ and $\sigma \in \mathbb{R}^+$. Then the image of the discrete Gaussians will usually not be a discrete Gaussian and our inference principle becomes nontrivial, yielding preference for one direction.

### 3.5 Empirical estimators

In practice, one does not have access to the precise distributions $p_X$ and $p_Y$, but only to a finite i.i.d. sample of observations $\{x_i, y_i\}_{i=1}^m$ drawn from the joint distribution $p_{X,Y}$. There are at least two natural ways of estimating $C_{X \to Y}$: using (8) and using (9).

For both cases, one needs to estimate the term $S(u) - S(v)$. We do this by simple preprocessing of the data: for the uniform reference measure (described in Subsection 3.1), we apply an affine transformation to the $x$-values such that $\min_{i=1}^m x_i = 0$ and $\max_{i=1}^m x_i = 1$, and similarly for the $y$-values. For the Gaussian reference measures (described in Subsection 3.2), we standardize the $x$-values by an affine transformation such that their mean becomes 0 and variance becomes 1, and similarly for the $y$-values. After this preprocessing step, the term $S(u) - S(v)$ vanishes, and we only need to consider the term $S(p_Y) - S(p_X)$.

For the estimation based on (8), we can simply plug in any entropy estimator for estimating the marginal entropies $S(p_X)$ and $S(p_Y)$. For example, in the one-dimensional case, we have used the following estimator [Kraskov et al., 2003]:

$$\hat{S}(p_X) := \psi(m) - \psi(1) + \frac{1}{m-1} \sum_{i=1}^{m-1} \log |x_{i+1} - x_i| \,, \quad (12)$$

where the $x$-values should be ordered ascendingly, i.e., $x_i \leq x_{i+1}$, and $\psi$ is the digamma function.[6]

For the alternative estimation based on (9), we observe that the integral can be estimated in the one-dimensional case by

$$\int \log |f'(x)| \, p(x) dx \approx \frac{1}{m-1} \sum_{i=1}^{m-1} \log \left| \frac{y_{i+1} - y_i}{x_{i+1} - x_i} \right| \,, \quad (13)$$

where we have assumed the $(x, y)$-pairs to be ordered according to $x_i \leq x_{i+1}$ (if either the numerator or the denominator of the fraction within the log is zero, that data point is not taken into account in the summation).

The difference between both estimators is subtle and not yet completely understood. In the deterministic case, the two estimators are identical (even for finite $m$). In the noisy case, however, (see also Section 4), one has to be careful with the second estimator (13), as it diverges as $m \to \infty$, and one has to compensate for this by considering the difference of this estimator and its analogon in the reverse direction (obtained by swapping the roles of $x$ and $y$).[7]

These estimators generalize straightforwardly to higher dimensions (i.e., the case described in Subsection 3.2). The preprocessing is then done by whitening the data (i.e., the data are linearly transformed such that they have the identity as covariance matrix). Because of space constraints, we omit the details here.

## 4 Adding small noise

In this section we provide a partial analysis of our method when small amounts of noise are present. Assuming that $X$ is the cause, and that the additive noise model holds, we bound the variance of the noise so that $S(p_Y)$ remains lower than $S(p_X)$.

Before presenting the result, let us define a measure of non-Gaussianity of a random variable $X \sim p_X$ with mean $\mu_X$ and variance $\sigma_X^2$ as

$$D_G(p_X) := D(p_X \,\|\, N(\mu_X, \sigma_X^2)). \quad (14)$$

---

[5]We emphasize that all derivations so far also hold for the discrete case if the integrals are read in a measure theoretic sense, then "densities" are probability mass functions (which are, at the same time, densities with respect to the counting measure).

[6]The digamma function is the logarithmic derivative of the gamma function: $\psi(x) = d/dx \log \Gamma(x)$. It behaves as $\log x$ asymptotically for $x \to \infty$.

[7]In the draft version of this paper, we only considered the estimator (13). In retrospect, it seems simpler to use entropy estimators instead. However, the entropy estimators perform empirically slightly worse, although it is not obvious whether that difference is statistically significant. For this reason, we decided to mention both estimators here. In the experiments section, we only report the results for the entropy estimator.

The Fisher information of $p_X$ is defined as

$$J(X) := \mathbb{E}\left(\frac{\partial \log p_X(x)}{\partial x}\right)^2. \quad (15)$$

**Lemma 2 (Noise-generated entropy)**
Suppose $X \sim p_X$, $X \perp\!\!\!\perp E$, with $\mathbb{E}(E) = 0$, $\mathrm{Var}(E) = 1$, and $X \perp\!\!\!\perp Z$ with $Z \sim N(0,1)$. If

$$D_G(p_{X+\sqrt{\sigma}E}) \geq D_G(p_{X+\sqrt{\sigma}Z}), \quad (16)$$

then

$$S(p_{X+\sqrt{\sigma}E}) \leq S(p_X) + \frac{1}{2}\log(\sigma J(X) + 1). \quad (17)$$

**Proof**: Integrating De Bruijn's identity

$$\frac{\partial}{\partial t} S(p_{X+\sqrt{t}Z}) = \frac{1}{2} J(X + \sqrt{t}Z),$$

we obtain

$$S(p_{X+\sqrt{\sigma}Z}) - S(p_X) = \frac{1}{2}\int_0^\sigma J(X + \sqrt{t}Z)dt.$$

Applying the convolution inequality for Fisher information

$$J(X + \sqrt{t}Z) \leq \frac{J(X)J(\sqrt{t}Z)}{J(X) + J(\sqrt{t}Z)}$$

to the above integral and substituting $J(\sqrt{t}Z) = \frac{1}{t}$ we obtain the statement of the lemma for Gaussian noise. On the other hand $D_G(p_X) = \frac{1}{2}\log(2\pi e \sigma_X^2) - S(p_X)$. Assuming (16), in the non-Gaussian case we have $S(p_{X+\sqrt{\sigma}E}) \leq S(p_{X+\sqrt{\sigma}Z})$. Note that when $X$ and $E$ both are Gaussian, inequality (17) becomes tight. □

This lemma implies the following theorem, which shows the robustness of our method.

**Theorem 1** *If $Y = f(X)$ and $S(p_Y) < S(p_X)$, for any $E$ satisfying inequality (16) with $\sigma < \frac{e^{2S(p_X) - 2S(p_Y)} - 1}{J(Y)}$ we have*

$$S(p_{Y+\sqrt{\sigma}E}) < S(p_X).$$

**Proof**: Since $S(p_X) - S(p_Y) > 0$ and $\sigma J(Y) < e^{2S(p_X) - 2S(p_Y)} - 1$ by Lemma 2 we obtain

$$S(p_X) > S(p_Y) + \frac{1}{2}\log(\sigma J(Y) + 1) \geq S(p_{Y+\sqrt{\sigma}E}). \quad \square$$

## 5 Experiments

In this section we describe some experiments which illustrate the theory above and show that our method can detect the true causal direction in many real-world data sets. Unless explicitly stated, we normalize the data to $[0,1]$ (i.e., we choose uniform reference measure) and use the empirical estimators $\hat{S}(p_X)$ (and $\hat{S}(p_Y)$) defined in (12). Complete source code for the experiments is provided online.

Table 1: The different input distributions $p_X$ and mechanisms $f$ used in the simulations

| | $p_X$ | | $f(x)$ |
|---|---|---|---|
| (A) | $U(0,1)$ | (a) | $x^{\frac{1}{3}}$ |
| (B) | $N(0,\sigma^2)$ | (b) | $x^{\frac{1}{2}}$ |
| (C) | $N(0.5,\sigma^2)$ | (c) | $x^2$ |
| (D) | $N(1,\sigma^2)$ | (d) | $x^3$ |
| (E) | $GM_{([0.3,0.7]^T,[\frac{\sigma}{2},\frac{\sigma}{2}]^T)}$ | (e) | $s_5(x)$ |

### 5.1 Simulated data

We consider some artificial data sets and investigate the performance of the proposed method in various simple deterministic and low noise situations. Let

$$Y = f(X) + \lambda E,$$

($\lambda \geq 0$) be an additive noise model, where the input $X$ has distribution $p_X$, $f$ is some monotonous function, the noise $E$ is independent of $X$ and is generated according to the distribution $p_E$.

Denote by $\phi(X|\mu,\sigma)$ a Gaussian pdf, by $GM_{\boldsymbol{\mu},\boldsymbol{\sigma}}$ a Gaussian mixture distribution with density

$$\rho(x) = \frac{1}{2}\big(\phi(x|\mu_1,\sigma_1) + \phi(x|\mu_2,\sigma_2)\big),$$

and let

$$s_n(x) = \sum_{i=1}^n \alpha_i \Phi(x|\mu_i,\sigma_i) \quad (18)$$

be a convex combination of Gaussian cdf's $\Phi(x|\mu_i,\sigma_i)$, with parameters $\alpha_i, \mu_i \in [0,1], \sigma_i \in [0,0.1]$.

In our experiments we investigate all the combinations of $p_X$, $p_E$, and $f$ described in Table 1. The input distributions were clipped to the interval $[0,1]$, and in each experiment the parameters of $s_5(x)$ are chosen randomly according to the uniform distribution. Figure 2 provides plots of the distributions and functions. The width parameter $\sigma$ in the densities was set to 0.2 in all the cases. For each setting, the experiments were repeated 100 times (sample size $m = 1000$), the results were averaged, and the percentage of correct inferences is shown in Table 2.

The experimental results suggest that in the deterministic case ($\lambda = 0$) the method fails when the regions of high derivative of $f$ coincide with input density peaks. Note that such coincidences necessarily lead to wrong decisions. However, if the derivative of the function strongly fluctuates and $p_X$ has many irregularly distributed peaks, it becomes increasingly unlikely that peaks of $p_X$ always meet peaks of $\log f'$. Hence, in the case of more structured $f$ like $s_5(x)$, independently of $p_X$, the true causal direction was inferred quite accurately with all considered input distributions.

Table 2: Empirical performance on artificial data

| $p_X \backslash f$ | (a) | (b) | (c) | (d) | (e) |
|---|---|---|---|---|---|
| $\lambda = 0$ | | | | | |
| (A) | 100 | 100 | 100 | 100 | 100 |
| (B) | 0 | 0 | 100 | 100 | 90 |
| (C) | 100 | 100 | 100 | 100 | 99 |
| (D) | 100 | 100 | 0 | 0 | 85 |
| (E) | 100 | 100 | 100 | 100 | 99 |
| $\lambda = 0.03, p_E = U(0,1)$ | | | | | |
| (A) | 100 | 100 | 100 | 100 | 100 |
| (B) | 0 | 0 | 100 | 100 | 86 |
| (C) | 100 | 100 | 100 | 100 | 97 |
| (D) | 100 | 100 | 0 | 0 | 86 |
| (E) | 100 | 100 | 97 | 100 | 100 |
| $\lambda = 0.03, p_E = N(0,1)$ | | | | | |
| (A) | 100 | 100 | 100 | 100 | 100 |
| (B) | 2 | 0 | 100 | 100 | 64 |
| (C) | 100 | 100 | 99 | 100 | 91 |
| (D) | 91 | 88 | 0 | 0 | 75 |
| (E) | 100 | 100 | 100 | 100 | 95 |
| $\lambda = 0.03, p_E = Lap(0,\sigma)$ | | | | | |
| (A) | 100 | 100 | 100 | 100 | 100 |
| (B) | 3 | 0 | 100 | 100 | 75 |
| (C) | 100 | 100 | 100 | 100 | 98 |
| (D) | 100 | 98 | 0 | 0 | 87 |
| (E) | 100 | 99 | 100 | 100 | 100 |

Table 3: Results for CauseEffectPairs data set (51 pairs): the percentage of the pairs for which a decision was made, and the fraction of correct decisions, for various methods.

| Method | Decisions (%) | Accuracy (%) |
|---|---|---|
| IGCI (uniform) | 100 | 78 |
| IGCI (Gaussian) | 100 | 76 |
| AN (Gaussian) | 20 | 100 |
| PNL | 82 | 95 |

work in the linear case, but adding only small fluctuations leads for reasonably small sample size ($\epsilon = 0.005$; $\omega = 40$; $X \sim N(0,1), N(0,\sigma^2), N(0.5,\sigma^2), N(1,\sigma^2), GM_{([0.3,0.7]^T,[\frac{\sigma}{2},\frac{\sigma}{2}]^T)}$; $m = 1000$) to a performance of about 90%.

### 5.2 Real data

**Cause-effect pairs**

We used the extended version of `CauseEffectPairs` from the task [Mooij and Janzing, 2010]. It consists of observations of 51 different pairs of variables from various domains, and the task for each pair is to find which is the cause and which the effect.[8] Note that most of the pairs in this data set have quite high noise levels. In Table 3 we provide comparative results of our method (using two different reference measures) and two other causal inference methods that are suitable for inferring the causal direction between pairs of variables: the Additive Noise (AN) model [Hoyer et al., 2009], using Gaussian Process Regression which assumes Gaussian noise, and an implementation of the Post-NonLinear (PNL) model [Zhang and Hyvärinen, 2009], using neural networks for constrained nonlinear independent component analysis. Both methods employ the HSIC independence test for accepting or rejecting the fitted models.

**German Rhine data**

The data consists of the water levels of the Rhine[9] measured at 22 different cities in Germany in 15 minute intervals from 1990 to 2008. It is natural to expect that there is a causal relationship between the water levels at the different locations, where "upstream" levels influence "downstream" levels.

We tested our method on all of the 231 pairs of cities. Since the measurements are actually time series, and

In the low noise regime ($\lambda = 0.03$) the performance remains similar for all $p_E$ listed in Table 1. However, higher noise can strongly affect the inferred direction, and its influence on the method's performance remains unclear. Since detailed investigation of the large noise regime is beyond the scope of this paper, we leave it for future work, mentioning only empirical results with real data (which are often rather noisy) in the next subsection.

To test the effect of small fluctuations we simulated data with different distributions and the functional relationship $Y = X + \epsilon \sin(\omega X)$. Our method should not the causal influence needs some time to propagate, we performed the experiments with shifted time series; for each pair, one series was shifted relatively to the other so as to maximize the correlation between them.

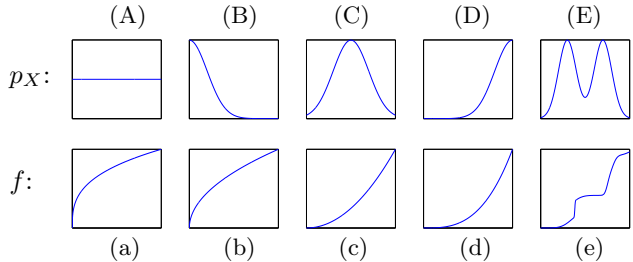

Figure 2: Input distributions $p_X$ (top row) and generating mechanisms $f$ (bottom row) corresponding to the 5 × 5 scenarios in Table 1.

---
[8]Accessible via http://webdav.tuebingen.mpg.de/cause-effect

[9]We are grateful to the German office "Wasser- und Schiffahrtsverwaltung des Bundes", which provides the data upon request.

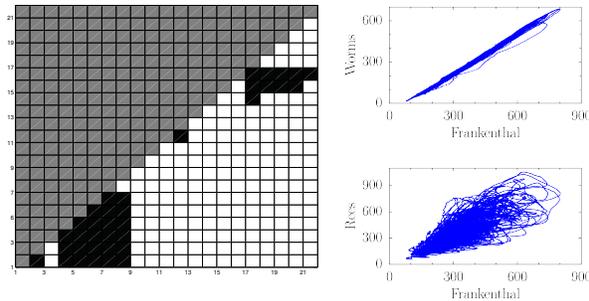

Figure 3: Results for the German Rhine data. All pairs out of in total 22 cities have been tested. White means a correct decision, black is a wrong decision, and the gray part can be ignored. On the right, typical data is illustrated for two measurement stations which are near to each other (top) and for two measurement stations farther apart (bottom), which shows that the noise increases significantly with the distance.

Figure 3 shows for each pair whether the decision is correct or not. It also shows some representative plots of the data. One can clearly see that the noise for two nearby cities is relatively low, but it can be quite large for two distant cities. Nevertheless, our method performed quite well in both situations: the overall accuracy, using the uniform reference measure, is 82% (189 correct decisions). The results for the Gaussian reference measure are similar (84%, 193 correct decisions).

## 6 Discussion

We have presented a method that is able to infer deterministic causal relations between variables with various domains. The assumption is that the distribution of the cause and the function satisfy a special kind of independence postulate (Postulate 2) with respect to some *a priori* chosen families of reference distributions. The choice of the families expresses a decision on how to split the description of the observed distribution into a part that contains the "essential" information and the one that merely contains information on the scaling and location. This way, it determines the notion of independence stated in our postulate. Empirically, both uniform distributions and Gaussians yielded good results.

We have shown that the method works on simulated data as well as real data. For causal relations that are given by a function plus small noise, we have also theoretically explained why the method still works.

The accuracy of the proposed method was shown to be competitive with existing methods. In terms of computation time, this method is orders of magnitude faster (in particular, it is linear in the number of data points). In addition, it can also handle the deterministic case, whereas existing methods only work in the presence of noise.

We would like to point out that in the very large noise regime, this method may completely fail; however, a better understanding of this case is needed. Moreover, one important line of our future research would be to find a way of estimating the confidence in the inferred causal direction.

### Acknowledgements

We are very grateful to Jan Lemeire for pointing out that deterministic relations are problematic for conventional causal inference methods. We thank Jonas Peters for helpful discussions.